\pdfoutput=1
\documentclass[11pt]{article}
\usepackage[table,usenames,dvipsnames]{xcolor}

\usepackage[]{acl}
\usepackage{tabularx}
\usepackage{comment}
\usepackage{placeins}
\usepackage{stfloats} 
\usepackage{placeins}
\usepackage{float} 
\usepackage{tabu} 
\usepackage{multirow}
\usepackage{caption}
\usepackage{booktabs}
\usepackage{times}
\usepackage{latexsym}
\usepackage{graphicx}
\usepackage{amsmath}
\usepackage{threeparttable}
\usepackage{enumitem}
\usepackage{booktabs}
\usepackage{amssymb}
\usepackage{hyperref}
\usepackage{subcaption}
\usepackage{pgf-pie}
\usepackage{pgfplots}
\pgfplotsset{compat=1.18}
\usepackage{pgfplotstable}

\usepackage{colortbl}
\usepackage{graphicx}
\usepackage{booktabs}

\usepackage{multirow}
\usepackage{booktabs}
\usepackage{array}
\usepackage{graphicx}
\usepackage{makecell}
\usepackage{enumitem}

\usepackage{tikz}
\usepackage{array}
\usepackage{tabularx}

\usepackage{titlesec}

\titlespacing*{\paragraph}{%
  0pt}{
  0.3\baselineskip}{
  1em}

\usepackage{times}
\usepackage{latexsym}
\usepackage{float} 
\usepackage[T1]{fontenc}

\usepackage[utf8]{inputenc}

\usepackage{microtype}

\usepackage{inconsolata}
\usepackage{caption}
\usepackage{pgfplots}
\pgfplotsset{width=10cm,compat=1.17}
\usepackage{pgf-pie}
\usepackage[most]{tcolorbox}
\usepackage{enumitem}
\usetikzlibrary{pgfplots.polar, positioning}
\usepgfplotslibrary{colormaps}

\newtcolorbox{mybox}[2][]{
  width=\textwidth,
  colback=white, 
  colframe=pastelpurplebox,
  fonttitle=\bfseries,
  coltitle=white, 
  title=#2,
  #1,
  colbacktitle=pastelpurplebox, 
  enhanced,
  attach boxed title to top left={yshift=-2mm, xshift=2mm},
  boxed title style={colframe=pastelpurple},
  separator sign={\ ---\ }
}

\newcolumntype{L}[1]{>{\raggedright\let\newline\\\arraybackslash\hspace{0pt}}m{#1}}
\newcolumntype{C}[1]{>{\centering\let\newline\\\arraybackslash\hspace{0pt}}m{#1}}
\newcolumntype{R}[1]{>{\raggedleft\let\newline\\\arraybackslash\hspace{0pt}}m{#1}}

\title{Personality Matters: User Traits Predict LLM Preferences in Multi-Turn \\ Collaborative Tasks}

\author{%
  Sarfaroz Yunusov$^\dagger$\textnormal{,} 
  Kaige Chen$^\dagger$\textnormal{,} Kazi Nishat Anwar$^\dagger$ \textnormal{and}  
  Ali Emami$^\ddagger$\\
  $^\dagger$Brock University, St. Catharines, Canada \\
  $^\ddagger$Emory University, Atlanta, USA \\
  \texttt{\{zw22fi, ax23ev, ss23gd\}@brocku.ca} \\
  \texttt{ali.emami@emory.edu} \\
}

\begin{document}
\maketitle

\begin{abstract}
As Large Language Models (LLMs) increasingly integrate into everyday workflows, where users shape outcomes through multi-turn collaboration, a critical question emerges: do users with different personality traits systematically prefer certain LLMs over 
others? We conduc-ted a study with 32 participants evenly distributed across four Keirsey personality types, evaluating their interactions with GPT-4 and Claude 3.5 across four collaborative tasks: data analysis, creative writing, information retrieval, and writing assistance. Results revealed significant personality-driven preferences: \textit{Rationals}  strongly preferred GPT-4, particularly for goal-oriented tasks, while \textit{idealists} favored Claude 3.5, especially for creative and analytical tasks. Other personality types showed task-dependent preferences. Sentiment analysis of qualitative feedback confirmed these patterns. Notably, aggregate helpfulness ratings were similar across models, showing how personality-based analysis reveals LLM differences that traditional evaluations miss.
\end{abstract}

\section{Introduction}
Large Language Models (LLMs) increasingly serve as collaborative tools in everyday workflows, supporting creative writing, code generation, and information retrieval \cite{Weisz_2021, gomez-rodriguez-williams-2023-confederacy, tong2024codejudgeevaluatingcodegeneration}. In multi-turn exchanges, users often refine their prompts or give feedback, causing the interaction to evolve in ways that reflect their individual style and traits. As a result, understanding user-specific factors that influence collaboration effectiveness becomes critical.


\begin{figure}[t]
    \centering
    \includegraphics[width=\columnwidth]{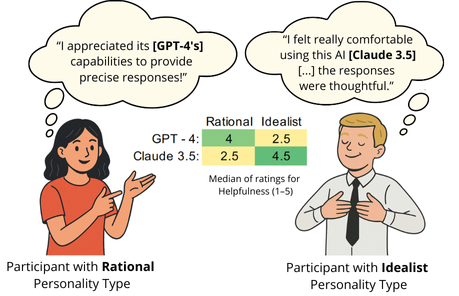}
\caption{Feedback from \textit{Rational} and \textit{Idealist} participants showing divergent preferences for models on the data analysis task.}
\end{figure} 

Recent studies suggest that LLMs exhibit distinct personality-like traits that affect user engagement \cite{serapiogarcía2025personalitytraitslargelanguage, hilliard2024elicitingpersonalitytraitslarge}, while human personality traits significantly influence trust in human-robot interactions \cite{Lim_2022}, LLM performance perceptions \cite{10580436}, and preferences in computer system interactions \cite{article}. However, most LLM evaluations remain limited to single-turn interactions \cite{mizrahi2024stateartmultipromptllm}, and the question of how users' personality traits influence their perception of different LLMs during extended collaborative tasks remains largely unexplored. Perceived helpfulness, how effectively users feel an LLM supports their specific needs and goals, is our measure to evaluate this personality-based influence. In this study, we investigate: \textbf{Do users with distinct personality traits find certain LLMs more helpful than others in multi-turn collaborative interactions?}

To address this question, we categorized participants using the Keirsey Temperament Sorter~\citep{keirsey1998please}, which condenses the Myers-Briggs Type Indicator ~\cite{myers1962mbti} into four temperament groups: \textit{Rationals} (analytical, strategic problem-solvers), \textit{Idealists} (imaginative, values-driven communicators), \textit{Guardians} (responsible, detail-oriented organizers), and \textit{Artisans} (adaptable, action-oriented creators)~\citep{keirsey1998please}.

We present a comprehensive framework for evaluating how personality traits influence perceived helpfulness in multi-turn human-LLM interactions across four task types, as illustrated in Figure~\ref{fig:main}. Our key contributions are:

\begin{figure*}[ht]
    \centering
    \includegraphics[width=1\textwidth]{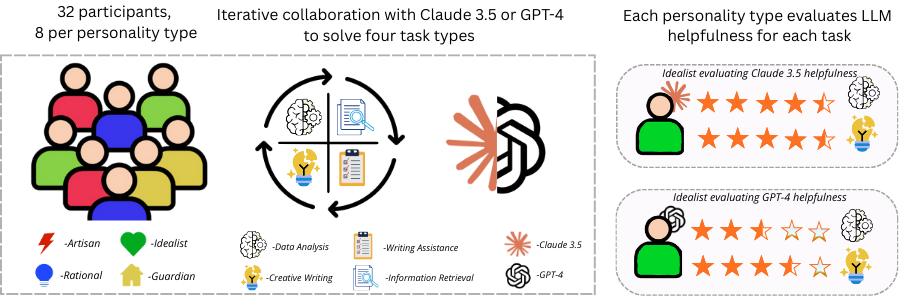}
    \caption{Overview of the experimental process, including participant grouping by personality type, collaboration with either GPT-4 or Claude 3.5 across task types, and subsequent helpfulness evaluations}
     
    \label{fig:main}
\end{figure*}
\begin{enumerate}[itemsep=1pt, topsep=1pt, parsep=0pt, leftmargin=*]

\item We design and implement four diverse task types that  simulate real-world collaborative uses of LLMs, balancing both open-ended and goal-oriented objectives. Participants engaged deeply with these tasks, using an average of 4.8 prompts and spending 11.2 minutes per task.

\item Through a controlled study with 32 participants, we reveal clear personality-driven preferences between two comparable LLMs, GPT-4 and Claude 3.5. \textit{Rationals} significantly favored GPT-4 over Claude 3.5, while \textit{Idealists} showed the opposite preference. Sentiment analysis of qualitative feedback further confirmed these divergent patterns.

\item We provide detailed qualitative analysis of participants' feedback, insights on effective prompting strategies, and demonstrate how aggregate helpfulness ratings mask important personality-dependent preferences.
\end{enumerate}

\section{Collaborative Task Design}

We designed four task types based on everyday use cases and participant-reported purposes for LLM usage (see Appendix Figure~\ref{fig:ai_purpose}). We balanced open-ended tasks, which allow for creative expression and subjective evaluation, with goal-oriented tasks, which have definitive correct answers.  Appendix Figures \ref{fig:problem_solving_description}-\ref{fig:email_description} provide detailed descriptions for the four tasks. Below are brief summaries of each task:

\paragraph{Task 1: Data Analysis – Student Loan Dataset}
Participants examined a synthetic dataset of 500 student loan records containing attributes such as age, gender, loan purpose, and account balances (sample in Appendix Table~\ref{tab:sample_dataset_problem_solving}). Working with the LLM, they needed to (1) determine the number of male students under 29 who applied for car loans, and (2) identify students with checking account balances below \$500. 

\paragraph{Task 2: Creative writing - Poem generation}
This exercise required participants to collaboratively generate two poems with the LLM: (1) a personalized poem about their relationship with their mother and (2) a short poem (8–10 lines) describing a fictional device that solves a problem in their chosen field. The participants selected their preferred version as their final submission.

\paragraph{Task 3: Writing Assistance – Email Refinement} 
For this task, participants collaborated with the LLM to improve a draft email, based on either a hypothetical or real situation. The goal was to maintain a personal tone while improving the clarity, tone, and structure of the message.

\paragraph{Task 4: Information Retrieval – Archaeological Investigation} 
This task required participants to evaluate and synthesize information from the LLM to solve a problem with a verifiable answer. In this specific instance, participants engaged in a simulated archaeological study about a lost city that disappeared between 1000 and 1500 AD. Given a set of clues, they worked with the LLM to determine the city's modern name, its historical region, and a plausible explanation for its disappearance.

\section{Experimental Setup}  

\paragraph{Participants and Recruitment}
We recruited 32 university students with diverse primary purposes of LLM usage and comfort, as detailed in Appendix Figure~\ref{fig:ai_usage}. Recruitment was conducted through university mailing lists and direct outreach. All participants provided their informed consent and the study received institutional ethics approval. A detailed breakdown of participant demographics, including education level, frequency of LLM use, and gender distribution, is available in Appendix Figure~\ref{fig:annotators_demongraphics}.

\paragraph{Personality Assessment}
Prior to the study, participants completed a self-assessment using the Keirsey Temperament Sorter\footnote{Self-assessment available at \href{https://www.keirsey.com}{keirsey.com}}, a tool widely used in occupational, educational, and counseling contexts to assess personality traits \cite{Robbins2020, berens2001quick}. Eight participants were selected from each of the four personality groups.

We selected the Keirsey Temperament Sorter over continuous trait models such as the Big Five due to its categorical clarity and practical feasibility. Keirsey assigns participants to one of four distinct personality types, while the Big Five describes personality along five continuous dimensions, making it challenging to form discrete and balanced groups without a substantially larger sample size. Although Keirsey may lack the psychometric precision of the Big Five, its simplicity and widespread use in applied settings like education and career development made it a practical fit for our study design.  




\begin{figure}[t!]
    \centering
    \includegraphics[width=190pt, height = 4.5cm]{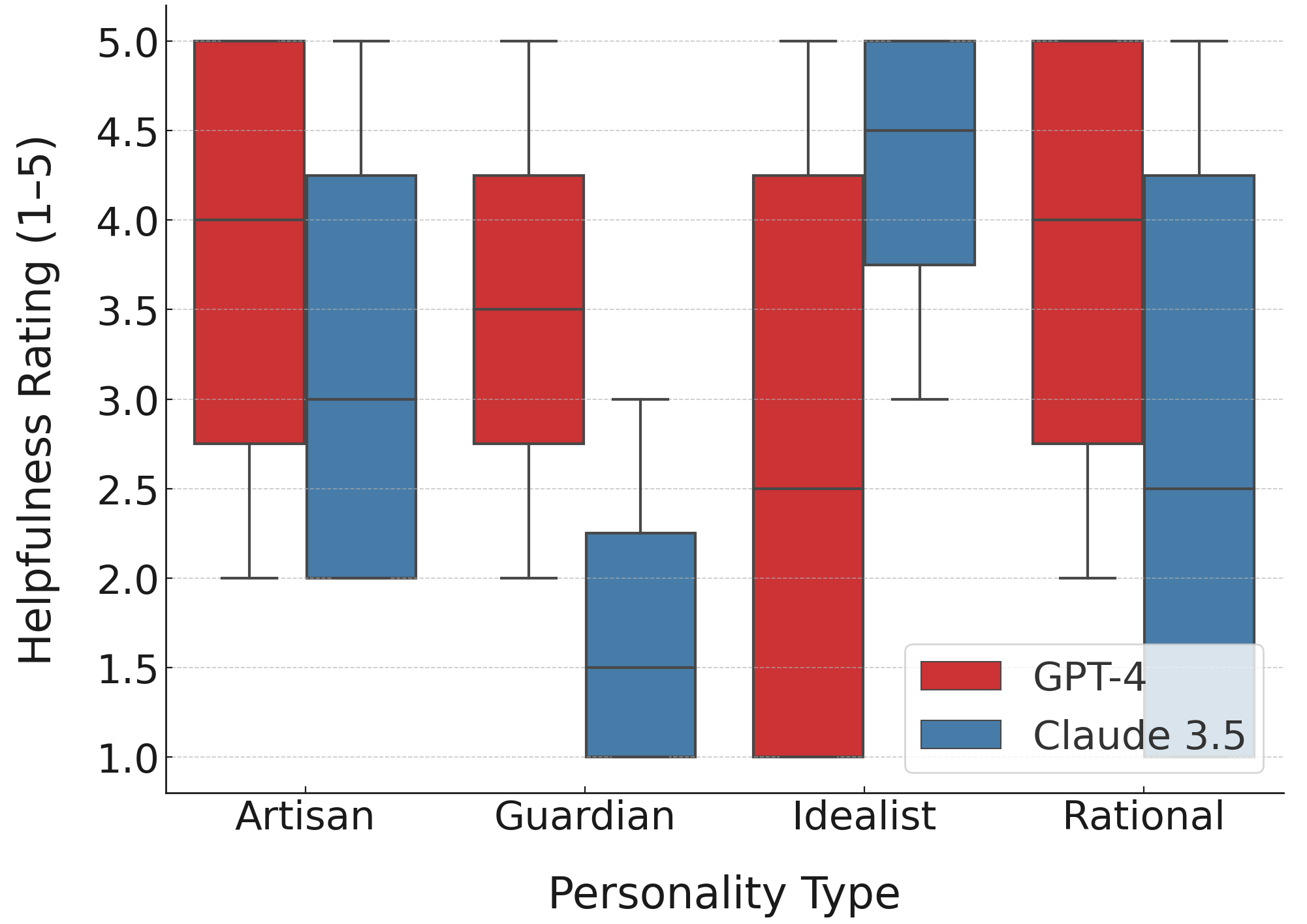}
     \vspace{-2mm}
\caption{Distribution of helpfulness ratings for the data analysis task by personality type and LLM. Boxes show interquartile range with median; whiskers show min/max values.}
    \label{fig:plot_box_problem_solving}
\end{figure}

\paragraph{Models}
The participants interacted with GPT-4 (\textit{ver. 0613}) \citep{openai2024gpt4technicalreport} or Claude 3.5 Sonnet \citep{anthropic2025claude3.5}, two models with comparable overall benchmark performance \citep{huang2024olympicarenamedalranksintelligent, monisha2025empiricalevaluationllmspredicting}\footnote{Using two models allowed us to maintain balance while keeping the participant pool size within resource budget.}. Each model was assigned to 16 participants, with an even distribution between the four personality groups (4 participants per personality type per model).

\paragraph{Experimental Procedure}  
The study was conducted in person under supervised conditions. Each participant completed all four tasks, presented in randomized order, using their assigned model. Participants engaged in multi-turn interactions without prior exposure to the tasks and were encouraged to interact as much as needed to complete each task. Upon completion of each task, participants completed a survey rating the model's helpfulness on a scale from 1 (not helpful) to 5 (extremely helpful), with the complete survey question provided in Appendix Figure~\ref{fig:question_after_task}. We chose helpfulness as our primary measure as it captures both objective task success and subjective user satisfaction in collaborative interactions. We defined \textit{helpfulness} as the extent to which the model effectively supported the participant in completing the assigned task based on their personal experience and satisfaction with the interaction.

A custom web application\footnote{Available at \url{https://personal-llm.ca}} was used for all interactions to standardize the experience and log of interaction data, including chat histories, number of prompts and task completion times. Importantly, this interface was identical regardless of the underlying model, ensuring participants were blind to which LLM they were using. This design prevented brand bias and preconceptions about specific models from influencing helpfulness ratings.


\begin{figure}[t!]
\centering
\begin{tikzpicture}
\begin{axis}[
    ylabel={Mean Helpfulness Rating},
    ylabel style={font=\fontsize{8pt}{8pt}\selectfont},
    xlabel style={font=\fontsize{10pt}{10pt}\selectfont},
    ybar,
    bar width=8pt,
    enlargelimits=0.2,
    xtick=data,
    xticklabels={Artisan, Guardian, Idealist*, Rational*},
    x tick label style={rotate=20,anchor=east, font=\scriptsize}, 
    yticklabel style={font=\footnotesize}, 
    legend style={at={(1.05,0.5)}, anchor=west, font=\scriptsize},
    grid=major,
    ymin=3, ymax=4.3, 
    width=5.5cm,
    height=4.4cm
]

\addplot+[ybar, fill=Blue, draw=black] coordinates {
    (1,3.625) (2,3.875) (3,3.625) (4,4.3125)
};
\addlegendentry{\rotatebox{90}{GPT-4}}

\addplot+[ybar, fill=Orange, draw=black] coordinates {
    (1,3.75) (2,3.625) (3,4.3125) (4,3.6875)
};
\addlegendentry{\rotatebox{90}{Claude 3.5}}

\end{axis}
\end{tikzpicture}
\vspace{-2mm}
\caption{
Mean helpfulness ratings of GPT-4 and Claude 3.5 by personality type across all tasks. Personality types marked with an asterisk (*) showed large effect sizes in model preference ($d \geq 0.8$), according to Cohen’s $d$.
}

\label{fig:llm_personality_ratings}
\end{figure}
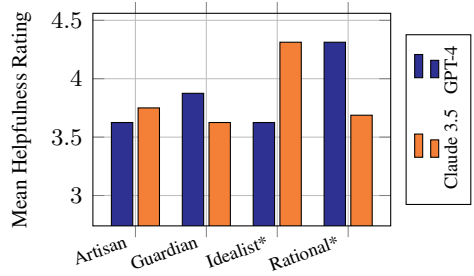

\section{Results} 
\paragraph{Personality Traits Shape Perceived Helpfulness in LLMs}
As shown in Figure~\ref{fig:llm_personality_ratings}, participants' personality traits significantly influenced their perceptions of LLM helpfulness. \textit{Idealists} consistently rated Claude 3.5 higher than GPT-4 on all tasks (4.4/5 versus 3.6/5 on average), while \textit{Rationals} showed the opposite pattern, strongly preferring GPT-4 (4.3/5 versus 3.7/5 for Claude), particularly for data analysis and information retrieval tasks (task-specific ratings available in Appendix Table~\ref{tab:mean_sd}). These preferences were especially pronounced in the data analysis task, as illustrated in Figure~\ref{fig:plot_box_problem_solving}, where Rationals rated GPT-4 up to 1.5 points higher than Claude 3.5. 

To assess the strength of these differences, we calculated Cohen's $d$, a widely accepted measure of effect size in behavioral and social sciences that expresses the difference between two group means in units of standard deviation \cite{cohen1988statistical}. Rationals showed a large effect ($d = 1.31$) favoring GPT-4, while Idealists showed a similarly large effect ($d = 1.44$) favoring Claude 3.5, where $d \geq 0.8$ typically indicates a large effect.

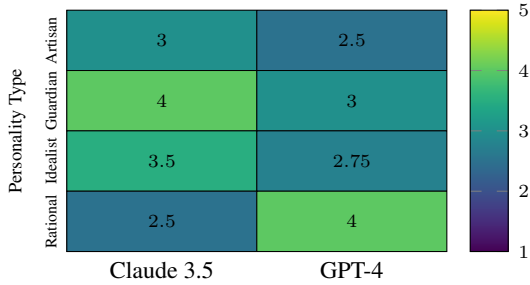
\begin{figure}[t]
    \centering
    \begin{tikzpicture}
        \begin{axis}[
            ylabel={\scriptsize Personality Type},
            enlargelimits=false,
            xtick=data,
            ytick=data,
            xticklabels={Claude 3.5, GPT-4},
            yticklabels={Artisan, Guardian, Idealist, Rational},
            y dir=reverse,
            colormap/viridis,
            colorbar,
            point meta min=1,
            point meta max=5,
            nodes near coords,
            nodes near coords align={center},
            every node near coord/.append style={font=\scriptsize},
            tick label style={font=\scriptsize},
            label style={font=\scriptsize},
            yticklabel style={font=\tiny, rotate=90},
            xticklabel style={font=\footnotesize},
            title style={font=\footnotesize},
            width=\columnwidth,
            height=5.5cm,
            scale only axis,
            x=2.5cm, 
            y=0.8cm    
        ]
        \addplot[
            matrix plot*,
            mesh/cols=2,
            point meta=explicit,
            line width=0.01pt,
            draw=black,
            nodes near coords,
            nodes near coords style={
                font=\scriptsize,
            },
            every node near coord/.append style={
                /pgf/number format/.cd,
                fixed,
                precision=2
            },
        ] table[meta=meta] {
            x y meta
            0 0 3.00
            1 0 2.50
            0 1 4.00
            1 1 3.00
            0 2 3.50
            1 2 2.75
            0 3 2.50
            1 3 4.00
        };
        \end{axis}
    \end{tikzpicture}
    \vspace{-2mm}
    \caption{Average sentiment score by personality type and LLM (where 5 indicates most positive sentiment)}
    \label{fig:sentiment_analysis_personality_llm_type}
\end{figure}

\begin{table}[t]
\footnotesize
\centering
\begin{tabular}{@{}p{1.6cm}l p{3.9cm} @{}}
\toprule
\shortstack{\scriptsize \textbf{Personality Type}} & \textbf{Model} & \textbf{Feedback} \\ \midrule
\shortstack{Rational \#1}  & \footnotesize GPT-4      & ``The chat was able to \textbf{complete all my requests,} including reading all excel data and provide solutions [...].''     \\
\shortstack{Idealist \#1}  & \footnotesize GPT-4      & ``It is particularly \textbf{frustrating} how poorly chatbots grasp how the human mind works [...].''     \\
\shortstack{Rational \#2}  & \footnotesize Claude 3.5 & ``\textbf{It's hard to rely} on this chatbot for things with any level of precision required [...].''  \\
\shortstack{Idealist \#2}  & \footnotesize Claude 3.5 & ``The accuracy of AI was \textbf{really good}. [...] at answering my question was \textbf{near perfect}.''  \\
\bottomrule
\end{tabular}
\vspace{-1mm}
\caption{Sample feedback from Rationals and Idealists toward GPT-4 and Claude 3.5}
\label{tab:personality_feedback_data_analysis}
\end{table}

Guardians and Artisans showed more task-dependent preferences. Guardians preferred GPT-4 for data analysis (3.5/5 vs. 1.75/5 for Claude 3.5) but favored Claude 3.5 for information retrieval (4.25/5 vs. 3.25/5 for GPT-4). Artisans showed only slight preferences across tasks, with a minor tendency to favor Claude 3.5 for writing assistance (see Appendix~\ref{sec:appendix_box_plots} for task-specific ratings).

\paragraph{Feedback Analysis Confirms Personality-Based Model Preferences} 
We analyzed participant feedback both quantitatively, using the \textit{BERT-Base-Multilingual-Uncased-Sentiment} \cite{nlptown2023sentiment} model, and qualitatively through manual review. Figure~\ref{fig:sentiment_analysis_personality_llm_type} shows that Rationals expressed more positive sentiment toward GPT-4, while Idealists and Guardians favored Claude 3.5. Table~\ref{tab:personality_feedback_data_analysis} provides examples of how feedback content reflected personality traits: Rationals evaluated based on task completion and precision, while Idealists focused on understanding and holistic accuracy. These suggest that personality influences \textit{both} model preferences \textit{and} the criteria used to assess helpfulness.

\section{Extended Analysis}

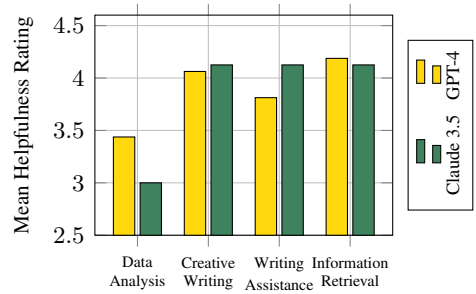
\begin{figure}[t]
\centering
\begin{tikzpicture}
\begin{axis}[
    ylabel={Mean Helpfulness Rating},
    ylabel style={font=\fontsize{8pt}{8pt}\selectfont},
    xlabel style={font=\fontsize{10pt}{10pt}\selectfont},
    ybar,
    bar width=8pt,
    enlargelimits=0.2,
    xtick=data,
    xticklabels={
        {\shortstack{Data\\Analysis}},
        {\shortstack{Creative\\Writing}},
        {\shortstack{Writing\\Assistance}},
        {\shortstack{Information\\Retrieval}}
    },
    x tick label style={rotate=0, font=\tiny},
    yticklabel style={font=\footnotesize}, 
    legend style={at={(1.05,0.5)}, anchor=west, font=\scriptsize},
    grid=major,
    ymin=2.8, ymax=4.3, 
    width=5.5cm,
    height=4.5cm
]

\addplot+[ybar, fill=Yellow!80!Orange, draw=black] coordinates {
    (1, 3.4375)
    (2, 4.0625)
    (3, 3.8125)
    (4, 4.1875)
};
\addlegendentry{\rotatebox{90}{GPT-4}}

\addplot+[ybar, fill=Green!60!Black, draw=black] coordinates {
    (1, 3)
    (2, 4.125)
    (3, 4.125)
    (4, 4.125)
};
\addlegendentry{\rotatebox{90}{Claude 3.5}}

\end{axis}
\end{tikzpicture}
\vspace{-2mm}
\caption{Mean helpfulness ratings of GPT-4 and Claude 3.5 across tasks}
\label{fig:all_tasks_helpfulness}
\end{figure}

\paragraph{Aggregated Ratings Hide Personality Differences}
Figure~\ref{fig:all_tasks_helpfulness} shows that when we aggregate helpfulness ratings across participants regardless of personality type, GPT-4 and Claude 3.5 receive nearly identical average ratings (3.87/5 vs. 3.85/5). This finding is notable as it demonstrates that traditional evaluation approaches that don't account for personality differences can mask significant variations in user experiences that only become visible through personality-stratified analysis.

\paragraph{Prompting Strategy Improves Task Success} We observed that successful data analysis task completion was associated with specific prompting strategies. Participants who provided concise prompts (averaging 122 words) were more likely to obtain correct responses than those who used verbose prompts (averaging 322 words). Longer prompts typically included large portions of the dataset pasted directly into the conversation, appearing to confuse models rather than help them (example shown in Appendix~\ref{sec:appendix_prompting_examples}).

\section{Related Work}  
Personality research has been explored in various frameworks, including the Myers-Briggs Type Indicator \cite{2013-29682-000}, and the Keirsey Temperament Sorter \cite{keirsey1998please}, both of which have been linked to language use and behavior \cite{Mehl2006-vo}. Our study builds on these theories to explore how personality traits influence human-LLM interactions. In the domain of Human-Computer Interaction, LLMs offer a new dimension for research, especially considering the non-deterministic nature of interactions between humans and models \cite{Chen_2024, etsenake2024understandinghumanllmdynamicliterature}. Four key phases, flow planning, facilitation, iteration, and testing, are central to understanding human-LLM exchanges \cite{10.1145/3613905.3650786}. Furthermore, recent work on personalized LLMs emphasizes how personality alignment between users and models can enhance engagement and trust \cite{andreas-2022-language, ouyang2022traininglanguagemodelsfollow, 
yunusov2024mirrorstoriesreflectingdiversitypersonalized,bubeck2023sparksartificialgeneralintelligence}.
Four key phases, flow-planning, facilitating, iterating, and testing, are central to understanding human-LLM exchanges \cite{10.1145/3613905.3650786}. 

\section{Conclusion}
This study demonstrates that personality traits significantly influence how users perceive LLM helpfulness during collaborative tasks. While aggregate ratings showed GPT-4 and Claude 3.5 as nearly equivalent, personality-stratified analysis revealed distinct patterns: Rationals strongly preferred GPT-4, particularly for goal-oriented tasks, while Idealists favored Claude 3.5, especially for creative and analytical work. These preferences were consistent across multiple measures—ratings, sentiment analysis, and qualitative feedback—with participants evaluating helpfulness through personality-specific criteria. As LLMs become standard tools in professional and personal workflows, understanding these personality-driven preferences becomes essential for both model development and deployment decisions. Future research should investigate whether these patterns hold across broader populations and explore how model adaptation to user personality might improve collaborative outcomes.

\section*{Limitations}
\paragraph{Demographic Scope} While our participants represented various genders, education levels, and degrees of familiarity with LLMs, they were primarily university students. Although expanding to a more demographically diverse sample would improve the generalizability of our findings to the broader population, it would require substantially greater recruitment efforts across multiple communities and potentially different incentive structures. Future work with more resources should address this limitation to determine whether the observed personality-based preferences hold across wider user groups.

\paragraph{Subjectivity in Ratings:} Helpfulness assessments are inherently subjective, influenced by participant expectations, previous LLM experience, and familiarity with tasks. This limitation is unavoidable in perception-based studies and represents a fundamental challenge in LLM evaluation research. Our multiple measures approach (combining ratings, sentiment analysis, and qualitative feedback) helps mitigate this limitation by triangulating findings across different assessment methods.

\paragraph{Model Selection:} Our focus on two high-performing LLMs (GPT-4 and Claude 3.5) was necessary for experimental feasibility. Each additional model would require doubling our participant pool to maintain a balanced assignment across all personality types (adding 32 more participants per model). This practical constraint limits insights into how personality traits might interact with a wider range of model capabilities, although the observed patterns between just two state-of-the-art models already reveal significant differences worth investigating further.

\paragraph{Personality Classification:} While the Keirsey Temperament Sorter provides a well-established framework for personality classification, personality exists on a spectrum rather than in discrete categories. Using a dimensional approach would require an impractically large sample size to adequately represent all possible combinations of trait levels. Our categorical approach allowed for feasible recruitment while still capturing meaningful personality-based differences.

\section*{Ethical Considerations}
We adhered to strict ethical standards throughout this research. The study received institutional ethics board approval after thorough review of our protocols, data handling procedures, and participant safeguards.

\paragraph{Informed Consent:} All participants provided explicit informed consent after receiving detailed information about the study purpose, tasks, data collection, and their rights. We specifically designed our consent process to ensure participants understood that their interactions would be analyzed for research purposes.

\paragraph{Data Privacy:} All participant data was anonymized immediately after collection, with personally identifiable information replaced by anonymous identifiers. Only demographic variables necessary for analysis were retained. The interaction logs and survey responses were secured on encrypted servers with access restricted to authorized researchers. This approach balances research needs with privacy protection.

\paragraph{Voluntary Participation:} Recruitment was carried out without coercion and compensation followed institutional guidelines at a level that recognized the time of the participants without creating unnecessary inducement. Participants were informed that they could withdraw at any time without penalty, ensuring genuine voluntary participation.


\bibliography{custom}

\clearpage
\onecolumn

\appendix
\section{Appendix}
\label{sec:appendix}

\begin{figure}[h]
    \centering
    \begin{subfigure}{0.32\textwidth}
        \centering
        \begin{tikzpicture}
        \begin{axis}[
            xlabel={LLM Usage Frequency},
            ylabel={Number of Participants},
            ylabel style={font=\fontsize{7pt}{11pt}\selectfont},
            xlabel style={font=\fontsize{7pt}{11pt}\selectfont},
            ybar,
            bar width=0.5cm,
            symbolic x coords={Often, Sometimes, Rarely, Never},
            xtick=data,
            xticklabel style={font=\scriptsize},
            yticklabel style={font=\scriptsize},
            x tick label style={rotate=25,anchor=east},
            grid=major,
            ymin=0, ymax=25,
            width=5.5cm,
            height=4.5cm,
            enlarge x limits=0.15,
            grid style=dashed,
            major grid style={line width=0.5pt,draw=gray!50},
        ]
        \addplot+[ybar, fill=Yellow!80!Orange, draw=black] coordinates {
            (Often,20) (Sometimes,11) (Rarely,1) (Never,0)
        };
        \end{axis}
        \end{tikzpicture}
        \caption{\small Distribution of participants by frequency of prior LLM usage}
    \end{subfigure}
    \hfill
        \begin{subfigure}{0.32\textwidth}
            \centering
            \begin{tikzpicture}
            \begin{axis}[
                xlabel={Education Level},         
                ylabel={Number of Participants},
                ylabel style={font=\fontsize{7pt}{11pt}\selectfont},
                xlabel style={font=\fontsize{7pt}{11pt}\selectfont},
                ybar,
                bar width=0.5cm,
                symbolic x coords={Bachelors, Masters, PhD},
                xtick=data,
                xticklabel style={font=\scriptsize},
                yticklabel style={font=\scriptsize},
                x tick label style={rotate=25,anchor=east},
                grid=major,
                ymin=0, ymax=20,
                width=5cm,
                height=4.5cm,
                enlarge x limits=0.4,
                grid style=dashed,
                major grid style={line width=0.5pt,draw=gray!50},
            ]
            \addplot+[ybar, fill=blue!80, draw=black] coordinates {
                (Bachelors,17) (Masters,14) (PhD,1)
            };
            \end{axis}
            \end{tikzpicture}
            \caption{\small Distribution of participants by education level}
        \end{subfigure}
    \hfill
    \begin{subfigure}{0.32\textwidth}
        \centering
        \begin{tikzpicture}
        \begin{axis}[
            xlabel={Self-Identified Gender},         
            ylabel={Number of Participants},
            ylabel style={font=\fontsize{7pt}{11pt}\selectfont},
            xlabel style={font=\fontsize{7pt}{11pt}\selectfont},
            ybar,
            bar width=0.5cm,
            symbolic x coords={Male, Female, Non-binary},
            xtick=data,
            xticklabel style={font=\scriptsize},
            yticklabel style={font=\scriptsize},
            x tick label style={rotate=25,anchor=east},
            grid=major,
            ymin=0, ymax=25,
            width=5 cm,
            height=4cm,
            enlarge x limits=0.2,
            grid style=dashed,
            major grid style={line width=0.5pt,draw=gray!50},
        ]
        \addplot+[ybar, fill=Green!60!Black, draw=black] coordinates {
            (Male,21) (Female,10) (Non-binary,1)
        };
        \end{axis}
        \end{tikzpicture}
        \caption{\small Distribution of participants by self-identified gender}
    \end{subfigure}
        \caption{Demographic distribution of participants across three dimensions: 
        (a) frequency of LLM usage, 
        (b) highest completed education level, and 
        (c) self-identified gender}
    \label{fig:annotators_demongraphics}
\end{figure}
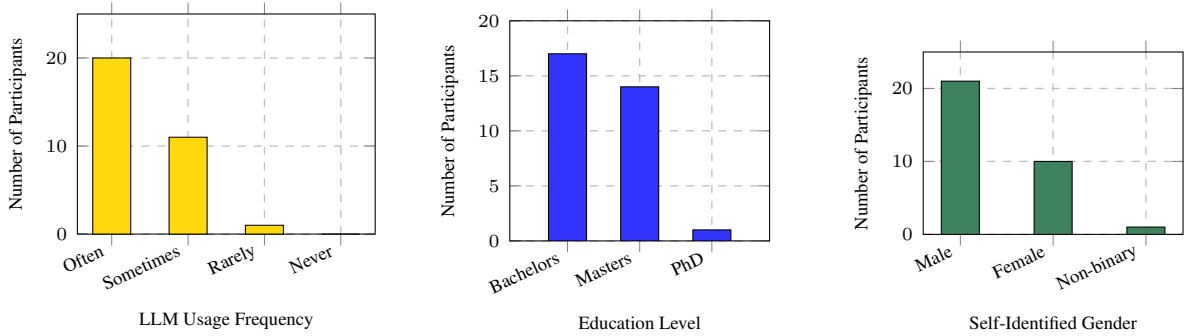

\vspace{0.5cm}

\begin{figure}[h]
    \centering
    \begin{subfigure}{0.48\textwidth}
        \centering
        \begin{tikzpicture}
            \begin{axis}[
                xlabel={Primary Use Case Category},                 ylabel={Percentage of Participants (\%)},
                ylabel style={font=\fontsize{7pt}{11pt}\selectfont},
                xlabel style={font=\fontsize{7pt}{11pt}\selectfont},
                ybar,
                bar width=0.6cm,
                symbolic x coords={Education, Creative Work, Entertainment, Research, Other},
                xtick=data,
                xticklabel style={font=\scriptsize},
                yticklabel style={font=\scriptsize},
                x tick label style={rotate=25,anchor=east},
                grid=major,
                ymin=0, ymax=100,
                width=6cm,
                height=5cm,
                enlarge x limits=0.2,
                grid style=dashed,
                major grid style={line width=0.5pt,draw=gray!50},
            ]
            \addplot+[ybar, fill=blue!80, draw=black] coordinates {
                (Education,90.60) 
                (Creative Work,68.80) 
                (Entertainment,15.60) 
                (Research,59.40) 
                (Other,3.10)
            };
            \end{axis}
        \end{tikzpicture}
        \caption{\small Primary use cases of LLMs reported by participants}
        \label{fig:ai_purpose}
    \end{subfigure}
    \hfill
    \begin{subfigure}{0.48\textwidth}
        \centering
        \begin{tikzpicture}
            \begin{axis}[
                xlabel={Comfort Level with LLMs (1 = Not Comfortable, 5 = Very Comfortable)},
                ylabel={Number of Participants},
                ylabel style={font=\fontsize{7pt}{11pt}\selectfont},
                xlabel style={font=\fontsize{7pt}{11pt}\selectfont},
                ybar,
                bar width=0.6cm,
                symbolic x coords={1, 2, 3, 4, 5},
                xtick=data,
                xticklabel style={font=\scriptsize},
                yticklabel style={font=\scriptsize},
                x tick label style={rotate=0,anchor=center},
                grid=major,
                ymin=0, ymax=20,
                width=6cm,
                height=5cm,
                enlarge x limits=0.2,
                grid style=dashed,
                major grid style={line width=0.5pt,draw=gray!50},
            ]
            \addplot+[ybar, fill=blue!80, draw=black] coordinates {
                (1,0) (2,0) (3,2) (4,13) (5,17)
            };
            \end{axis}
        \end{tikzpicture}
        \caption{\small Participant comfort level with using LLMs}
        \label{fig:ai_comfort}
    \end{subfigure}
    \caption{Participants' report on LLM usage patterns: (a) primary use cases (b) comfort levels}
    \label{fig:ai_usage}
\end{figure}
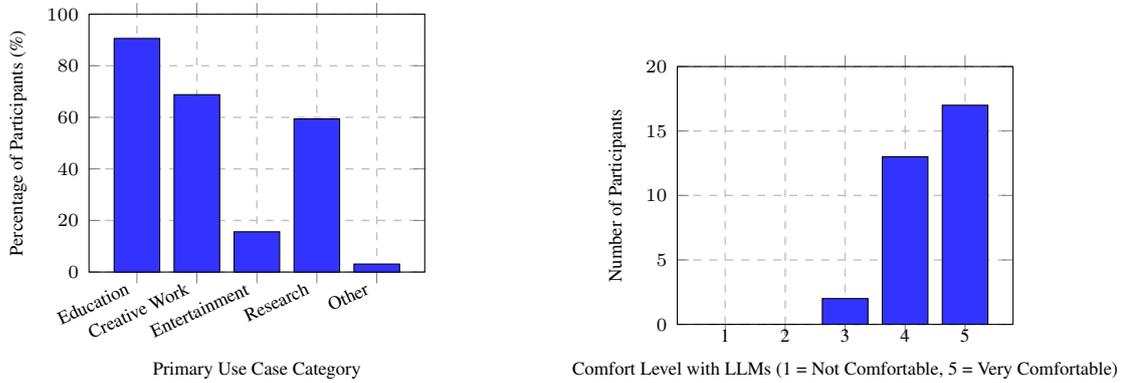

\vspace{1cm}

\begin{table}[h]
    \centering
    \small
    \renewcommand{\arraystretch}{1.2} 
    \rowcolors{2}{gray!15}{white} 
    \begin{tabularx}{0.75\textwidth}{X c c c c} 
        \toprule
        \textbf{Loan Purpose} & \textbf{Checking (\$)} & \textbf{Savings (\$)} & \textbf{Gender} & \textbf{Age} \\ 
        \midrule
        Business & 63 & 45 & F & 29 \\
        Car & 10 & 0 & M & 39 \\
        Car & 54 & 874 & F & 24 \\
        Car & 0 & 0 & F & 28 \\
        Education & 21 & 1000 & M & 19 \\
        Education & 400 & 450 & M & 39 \\
        Furniture & 0 & 100 & M & 48 \\
        \bottomrule
    \end{tabularx}
    \caption{Sample dataset used in the data analysis task}
    \label{tab:sample_dataset_problem_solving}
\end{table}

\begin{figure*}[t]
    \centering
    \includegraphics[width=300pt]{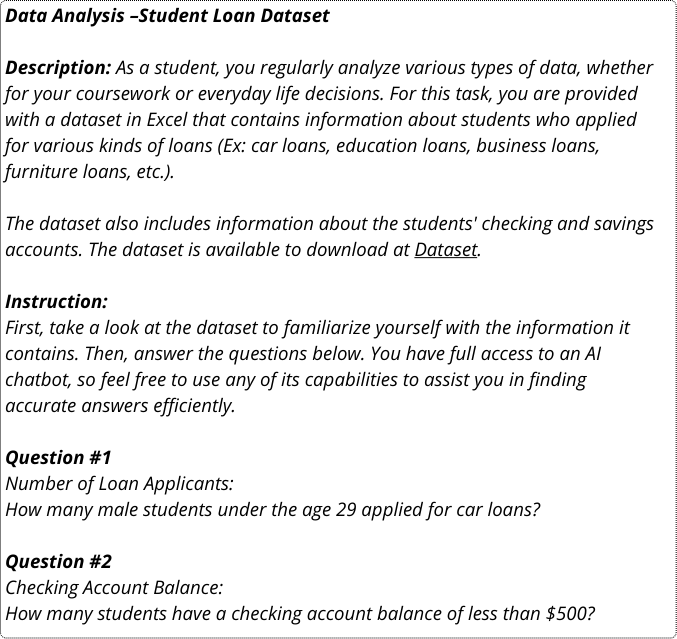}
    \caption{Description of data analysis task}
\label{fig:problem_solving_description}
\end{figure*}

\begin{figure*}[ht]
    \centering
    \includegraphics[width=300pt]{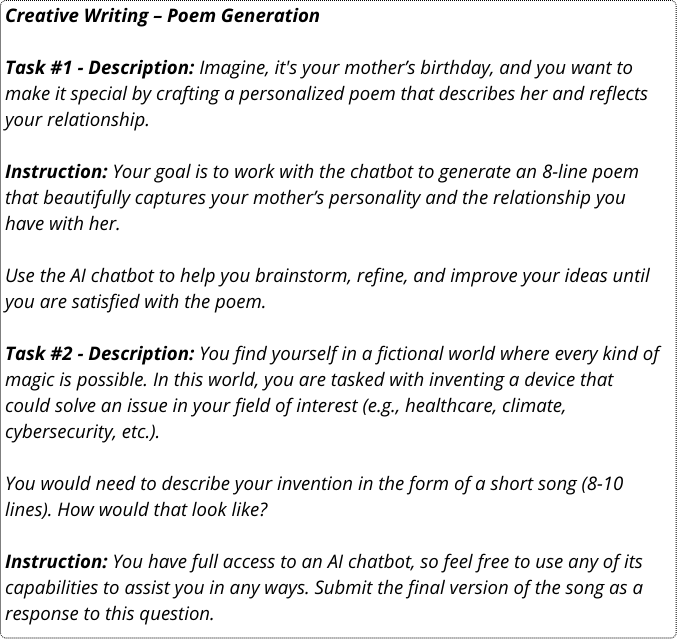}
    \caption{Description of creative writing task}
\label{fig:creative_description}
\end{figure*}

\begin{figure*}[ht]
    \centering
    \includegraphics[width=390pt]{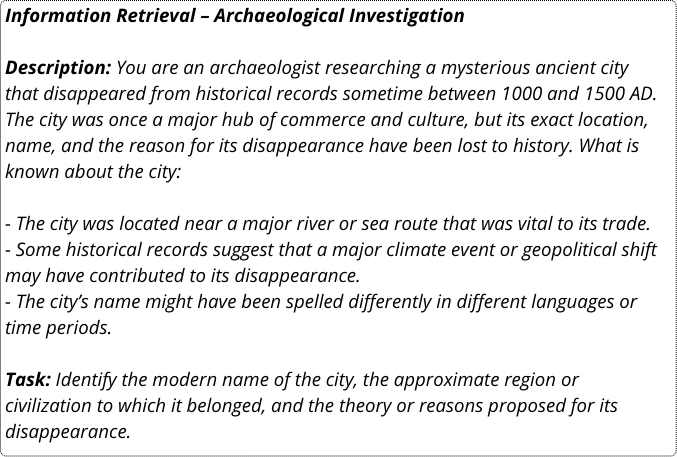}
    \caption{Description of information retrieval task}
\label{fig:info_description}
\end{figure*}

\begin{figure*}[ht]
    \centering
    \includegraphics[width=390pt]{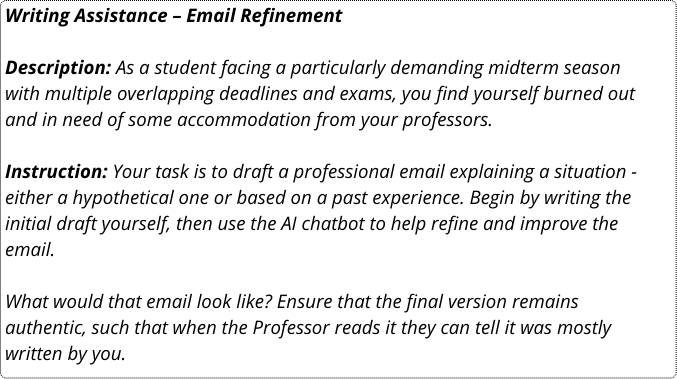}
    \caption{Description of writing assistance task}
\label{fig:email_description}
\end{figure*}

\clearpage 
\begin{figure*}[p]
    \centering
    \includegraphics[width=390pt]{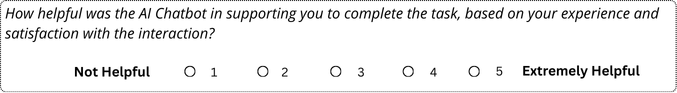}
    \caption{Helpfulness assessment question completed by participants after each task}
    \label{fig:question_after_task}
\end{figure*}

\begin{table}[t]
\centering
\begin{tabular}{c c c cccc}
\toprule
\makecell{\textbf{Personality} \\ \textbf{Type}} & 
\makecell{\textbf{Model}} & 
\makecell{\textbf{Data} \\ \textbf{Analysis}} & 
\makecell{\textbf{Creative} \\ \textbf{Writing}} & 
\makecell{\textbf{Writing} \\ \textbf{Assistance}} & 
\makecell{\textbf{Information} \\ \textbf{Retrieval}} \\
\midrule

\multirow{2}{*}{Artisan}
& \multirow{1}{*}{\footnotesize GPT-4} & \small 3.75 $\pm$ 1.50 & \small 3.75 $\pm$ 1.89 & \small 3.25 $\pm$ 1.20 & \small 4 $\pm$ 1.41 \\
& \multirow{1}{*}{\footnotesize Claude 3.5} & \small 3.25 $\pm$ 1.20 & \small 3.5 $\pm$ 0.58 & \small 3.5 $\pm$ 0.50 & \small 4 $\pm$ 1.00 \\
\midrule

\multirow{2}{*}{Guardian}
& \multirow{1}{*}{\footnotesize GPT-4} & \small 3.5 $\pm$ 1.29 & \small 4.25 $\pm$ 0.95 & \small 4.25 $\pm$ 0.57 & \small 3.25 $\pm$ 1.29 \\
& \multirow{1}{*}{\footnotesize Claude 3.5} & \small 1.75 $\pm$ 0.95 & \small 4.25 $\pm$ 0.95 & \small 4.25 $\pm$ 0.95 & \small 4.25 $\pm$ 0.95 \\

\midrule

\multirow{2}{*}{Idealist}
& \multirow{1}{*}{\footnotesize GPT-4} & \small 2.75 $\pm$ 1.00 & \small 3.75 $\pm$ 0.50 & \small 3.5 $\pm$ 1.29 & \small 4.5 $\pm$ 1.00 \\
& \multirow{1}{*}{\footnotesize Claude 3.5} & \small 4.25 $\pm$ 0.95 & \small 4.5 $\pm$ 0.57 & \small 4.25 $\pm$ 0.50 & \small 4.5 $\pm$ 0.95 \\

\midrule

\multirow{2}{*}{Rational}
& \multirow{1}{*}{\footnotesize GPT-4} & \small 3.75 $\pm$ 1.50 & \small 4.5 $\pm$ 0.57 & \small 4 $\pm$ 0.81 & \small 5 $\pm$ 0.00 \\
& \multirow{1}{*}{\footnotesize Claude 3.5} & \small 2.75 $\pm$ 2.30 & \small 4 $\pm$ 1.50 & \small 4 $\pm$ 0.50 & \small 3.5 $\pm$ 1.29 \\
\bottomrule
\end{tabular}
\caption{Average helpfulness ratings (1–5 scale) for GPT-4 and Claude 3.5 across personality types and task categories. Values represent mean ± standard deviation.}
\label{tab:mean_sd}
\end{table}

\clearpage
\subsection{Task-Based Helpfulness Ratings by Personality Types}
\label{sec:appendix_box_plots}
\vspace{-1cm}

\begin{figure}[ht]
    \centering
    Data Analysis
    
    \vspace{0.5em} 
    \includegraphics[width=190pt, height=4.5cm]{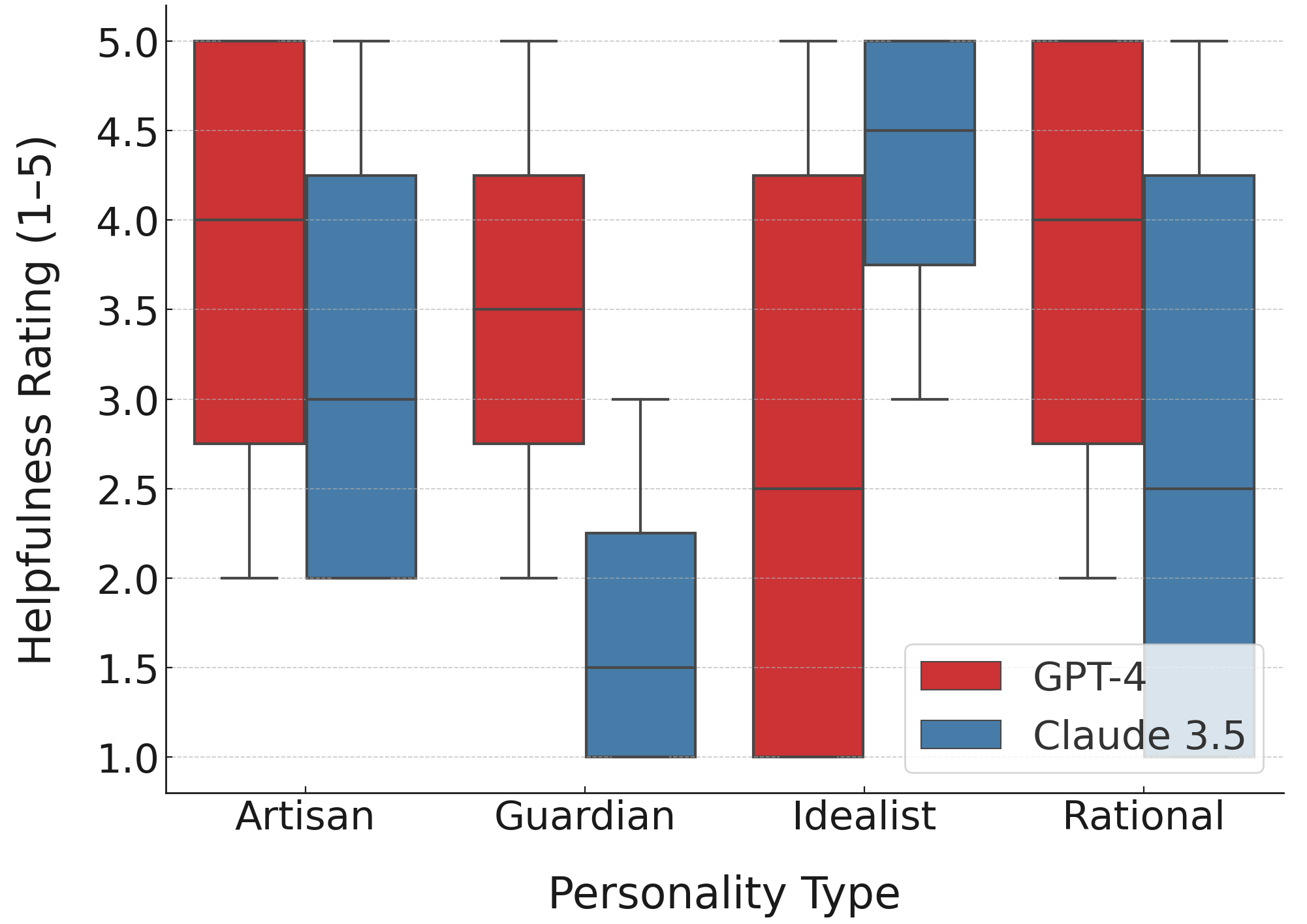}

    \caption{Distribution of helpfulness ratings for the data analysis task by personality type and LLM}
    \label{fig:plot_box_problem_solving_2}
\end{figure}

\begin{figure*}[ht]
    \centering
    Creative Writing
    
    \vspace{0.5em} 
    \includegraphics[width=190pt, height = 4.5cm]{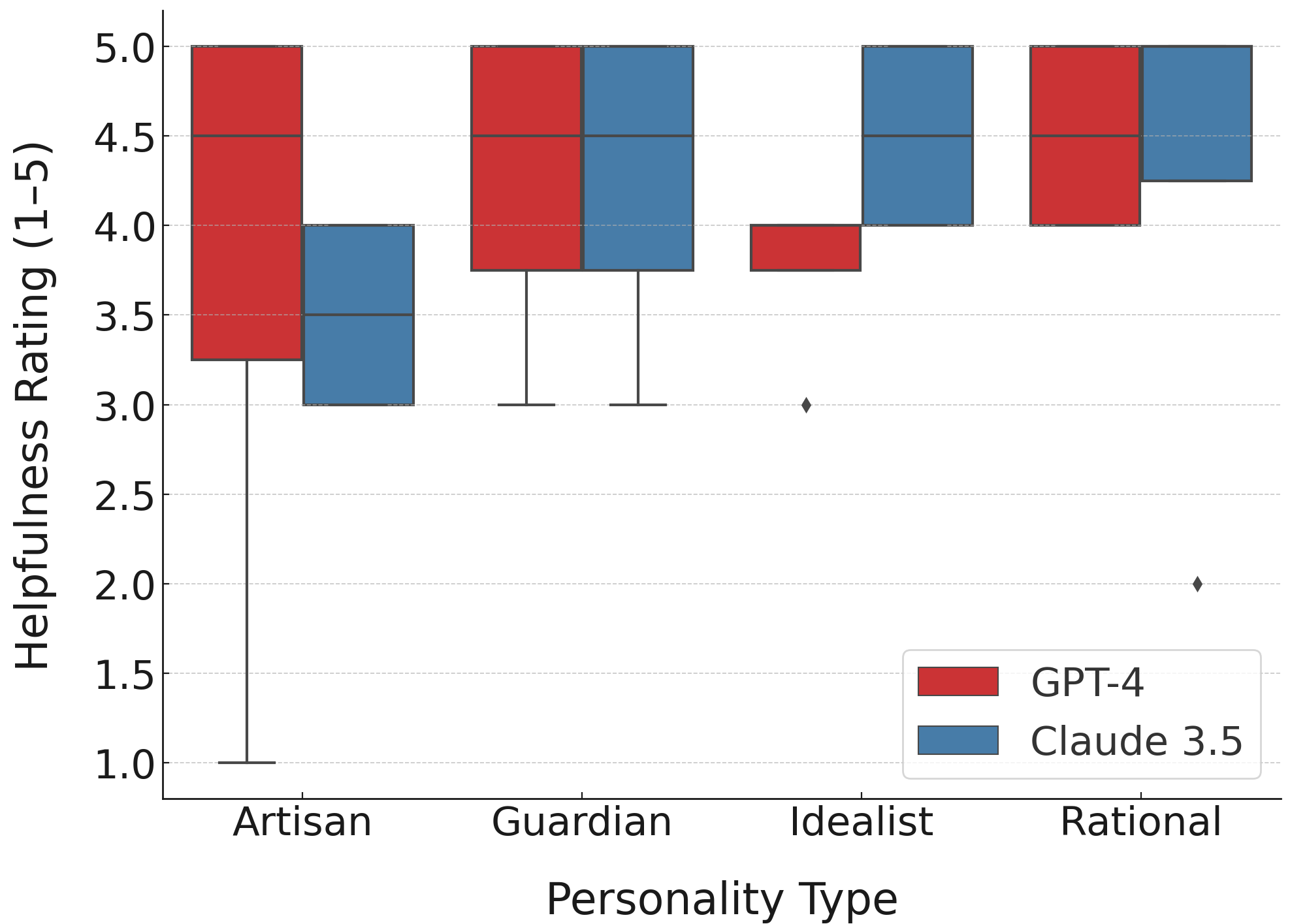}
    \caption{Distribution of helpfulness ratings for creative writing task by personality type and LLM}
    \label{fig:box_plot_creative}
\end{figure*}

\begin{figure*}[ht]
    \centering
    Writing Assistance
    
    \vspace{0.5em} 
    \includegraphics[width=190pt, height = 4.5cm]{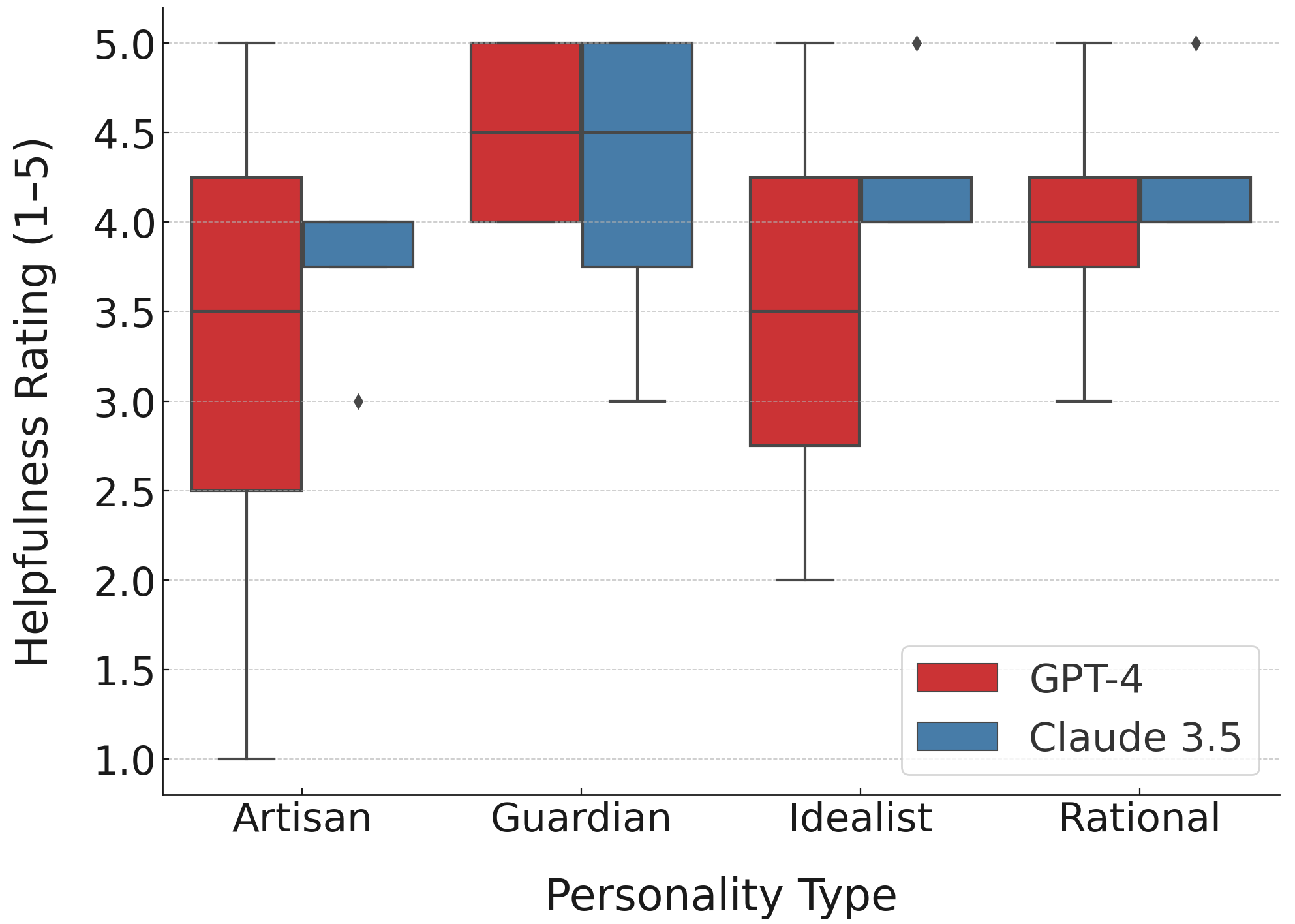}
    \caption{Distribution of helpfulness ratings for writing assistance task by personality type and LLM}
    \label{fig:box_plot_assistance}
\end{figure*}

\begin{figure*}[ht]
    \centering
    Information retrieval
    
    \vspace{0.5em} 
    \includegraphics[width=190pt, height = 4.5cm]{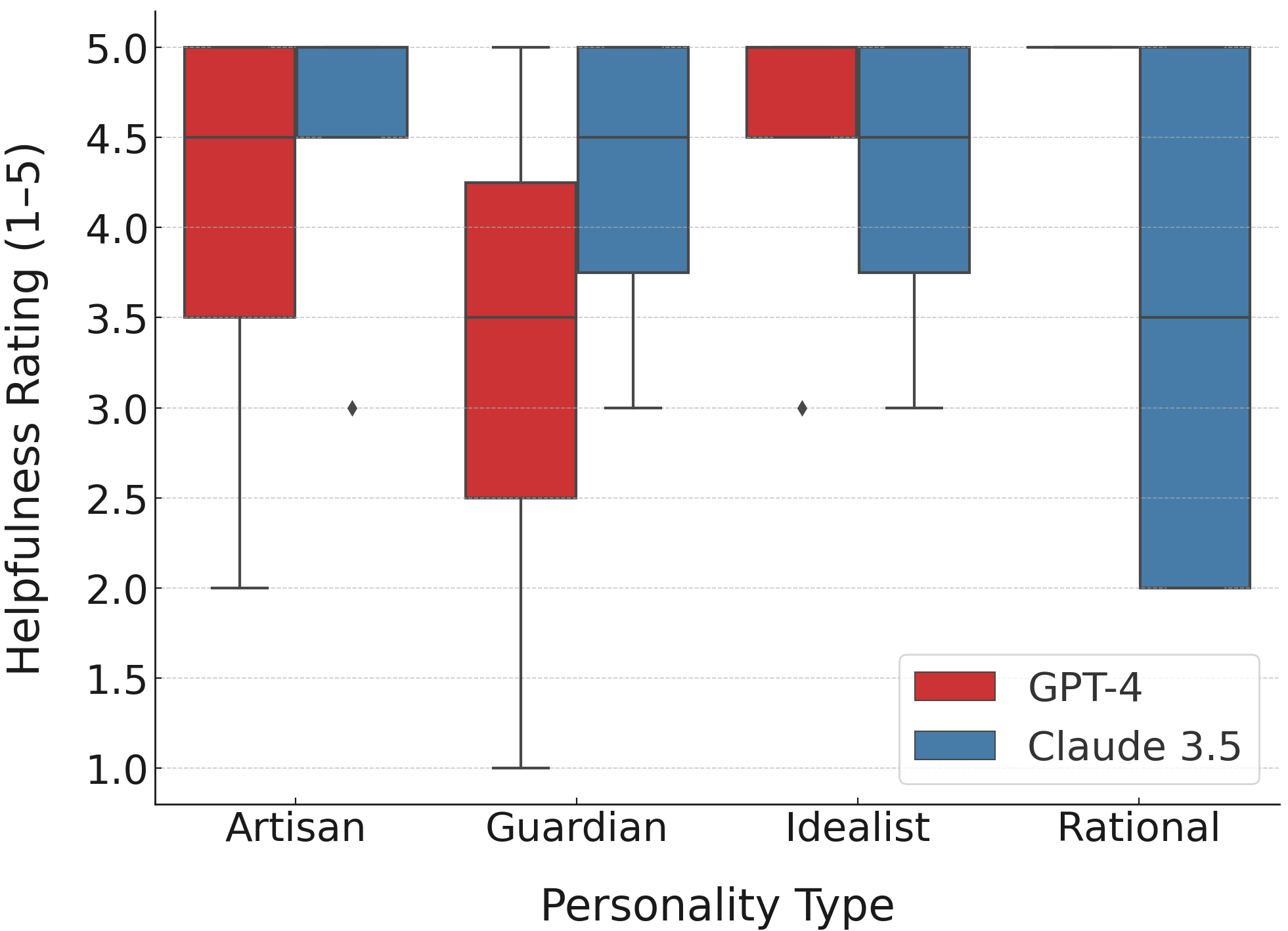}
    \caption{Distribution of helpfulness ratings for information retrieval task by personality type and LLM}
    \label{fig:box_plot_Info}
\end{figure*}

\subsection{Examples of Prompting Strategy and Data Analysis Task Outcomes}
\label{sec:appendix_prompting_examples}
\vspace{-3cm}

\begin{tcolorbox}[
    colback=blue!1,
    colframe=blue!40!black,
    title=Participant A (Correct Outcome),
    fonttitle=\bfseries,
    before title=\vspace{1.5mm},
    after title=\vspace{1.5mm},
    sharp corners=south,
    boxrule=0.6pt
]
\textbf{Prompt:}  
I have a dataset in excel and I need to extract some data out of it. I have the following columms: Loan Purpose Checking Savings Gender Age Credit Risk

\vspace{0.5em}
\textbf{Response:}  
That's great. I'd be happy to help you extract data. Could you please provide more specific details on what exactly you're looking to extract from these columns? For example, are you looking for average age of people with a certain credit risk or anything in specific? The more details you provide, the better I can guide you.

\vspace{0.5em}
\textbf{Prompt:}  
I have the sheet opened up on google sheets. Here is the first question. I need to find how many male students under the age 29 applied for car loans. How can i do that?

\vspace{0.5em}
\textbf{Response:}  
Sure, you can use the `COUNTIFS` function in Google Sheets to achieve this. The `COUNTIFS` function is used to apply multiple conditions on multiple ranges. 

Considering the columns are as follows:

- Loan Purpose = A

- Gender = B

- Age = C

Use the following formula:

`=COUNTIFS(A:A, "Car Loan", B:B, "Male", C:C, "<29")` [...] As a result, the function will return the count of male students under the age of 29 who applied for car loans. 

\vspace{0.5em}
\textbf{Prompt:}  
Just output the sample of formula I should use

\vspace{0.5em}
\textbf{Response:}  
Certainly! Here's the specific formula:
`=COUNTIFS(A:A, "Car Loan", B:B, "Male", C:C, "<29")` This function should provide the number of male students under age 29 who applied for car loans. As previously identified, you'll replace A:A, B:B, and C:C with the columns in your Google Sheet where Loan Purpose, Gender, and Age data are respectively. [...]

\end{tcolorbox}

\vspace{1em}

\begin{tcolorbox}[
    colback=red!1,
    colframe=red!40!black,
    title=Participant B (Incorrect Outcome),
    fonttitle=\bfseries,
    before title=\vspace{1.5mm},
    after title=\vspace{1.5mm},
    sharp corners=south,
    boxrule=0.6pt
]
\textbf{Prompt:}  
I am giving you a series of data about loan approvals and I want you to give me some information about it. Understood? \\[0.2em]
Dataset: 

Loan Purpose	Checking 	Savings	Gender	Age	Credit Risk

Business	\$0	\$739	M	23	Low

Furniture	\$0	\$1,230	M	32	High

[...]

[...]

Furniture	\$457	\$318	M	40	Low

Business	\$5,133	\$698	M	36	High

Car	\$0	\$33	M	29	Low

\vspace{0.5em}
\textbf{Response:}  
This dataset is quite extensive, thus providing specific information might need a more particular request. However, I can provide some general insights based on the given data:

1. Loan Purposes: The data covers different purposes for loan applications such as Business, Furniture, Car, Technology, and Education. Business loans appear most frequent while Education, Furniture, and Car also have a significant presence.

[...]

5. Credit Risk: The dataset contains information about Low and High credit risk individuals, with both categories appearing quite frequently.
For more detailed analysis or specific information, please make a precise request or provide me with the exact parameters for analysis.

\vspace{0.5em}
\textbf{Prompt:}  
Prompt: All I need you to answer is:

1. How many male students, under the age of 19, applied for car loans?

2. How many students, have a checking account balance, of less than \$500?

\vspace{0.5em}
\textbf{Response:}  
Certainly! I'll analyze the dataset to answer your specific questions.
To answer this, I'll count the entries that meet all of these criteria:

- Gender: M (Male)

- Age: Under 29

- Loan Purpose: Car

After going through the data, I found 9 entries that meet these criteria. \\[0.2em]
2. How many students have a checking account balance of less than \$500?

For this question, I'll count the entries where:

- Loan Purpose: Education

- Checking account balance: Less than \$500 \\[0.2em]
After analyzing the data, I found 16 entries that meet these criteria.\\[0.2em]
It's important to note that these counts are based on the data provided in the dataset. If the dataset is a sample or if there's any missing data, the actual numbers in the full population might be different.
\end{tcolorbox}

\vspace{0.1em}
\captionof{figure}{Interaction samples from two participants: concise prompting by Participant A led to a correct result, while verbose input from Participant B resulted in model confusion and an incorrect answer}
\label{fig:promting_sample}

\end{document}